\documentclass[letterpaper, 10 pt, conference]{ieeeconf}  

\IEEEoverridecommandlockouts                              

\usepackage{graphics} 
\usepackage{epsfig} 
\usepackage{mathptmx} 
\usepackage{times} 
\usepackage{amsmath} 
\usepackage{amssymb}  

\usepackage{bbding}

\usepackage[utf8]{inputenc} 
\usepackage[T1]{fontenc}    
\usepackage{hyperref}       
\usepackage{url}            
\usepackage{booktabs}       
\usepackage{amsfonts}       
\usepackage{nicefrac}       
\usepackage{microtype}      
\usepackage{xcolor}         
\usepackage{cleveref}
\usepackage{graphicx}
\usepackage{float}
\usepackage{multirow}
\usepackage{diagbox}
\usepackage{booktabs}

\DeclareMathOperator*{\argmin}{arg\,min}
\usepackage{algorithm}
\usepackage{algpseudocode}
\usepackage{arydshln}

\usepackage{booktabs}

\usepackage{multirow}
\usepackage{listings}
\definecolor{dkgreen}{rgb}{0,0.6,0}
\definecolor{gray}{rgb}{0.5,0.5,0.5}
\definecolor{mauve}{rgb}{0.58,0,0.82}
\title{\LARGE \bf
End-to-End Affordance Learning for Robotic Manipulation
}

\author{Yiran Geng$^{*1}$, Boshi An$^{*1}$, Haoran Geng$^{1}$, Yuanpei Chen$^{2}$, Yaodong Yang$^{\dagger2}$, Hao Dong$^{\dagger1}$
\thanks{$1$ CFCS, School of CS, Peking University}
\thanks{$2$ Institute for AI, Peking University \& BIGAI}
\thanks{The first two authors contributed equally. }
\thanks{Corresponding to \{hao.dong, yaodong.yang\}@pku.edu.cn}
}

\begin{document}

\maketitle

\begin{abstract}


Learning to manipulate 3D objects in an interactive environment has been a challenging problem in Reinforcement Learning (RL). In particular, it is hard to train a policy that can generalize over objects with different semantic categories, diverse shape geometry and versatile functionality. 
Recently, the technique of visual affordance has shown great prospects in providing object-centric information priors with effective actionable semantics. 
As such, an effective policy can be trained to open a door by knowing how to exert force on the handle. However, to learn the affordance, it often requires human-defined action primitives, which limits the range of applicable tasks. 
In this study, we take advantage of visual affordance by using the contact information generated during the RL training process to predict contact maps of interest. Such contact prediction process then leads to an end-to-end affordance learning framework that can generalize over different types of manipulation tasks. Surprisingly, the  effectiveness of such framework holds even under the multi-stage and the multi-agent scenarios. 
We tested our method on eight types of manipulation tasks.  
Results showed that our methods 
outperform baseline algorithms, including visual-based affordance methods and RL methods, by a large margin on the success rate. 
The demonstration can be found at \url{https://sites.google.com/view/rlafford/}.


\end{abstract}

\section{INTRODUCTION}

Learning to manipulate objects is a fundamental problem in RL and robotics.
An end-to-end learning approach can explore the reach range of future intelligent robotics. 
Recently, researchers have shown an increased interest in visual affordance~\cite{https://doi.org/10.48550/arxiv.2203.00352, mandikal2020graff,Mo_2021_ICCV, wang2021adaafford, wu2022vatmart}, \emph{i.e.}, a task-specific prior representation of objects. Such representations provide the agents with semantic information of objects, allowing better performance of manipulation.

\begin{figure}[h]
    \centering
    \includegraphics[trim=150 155 265 80,clip, scale=0.7]{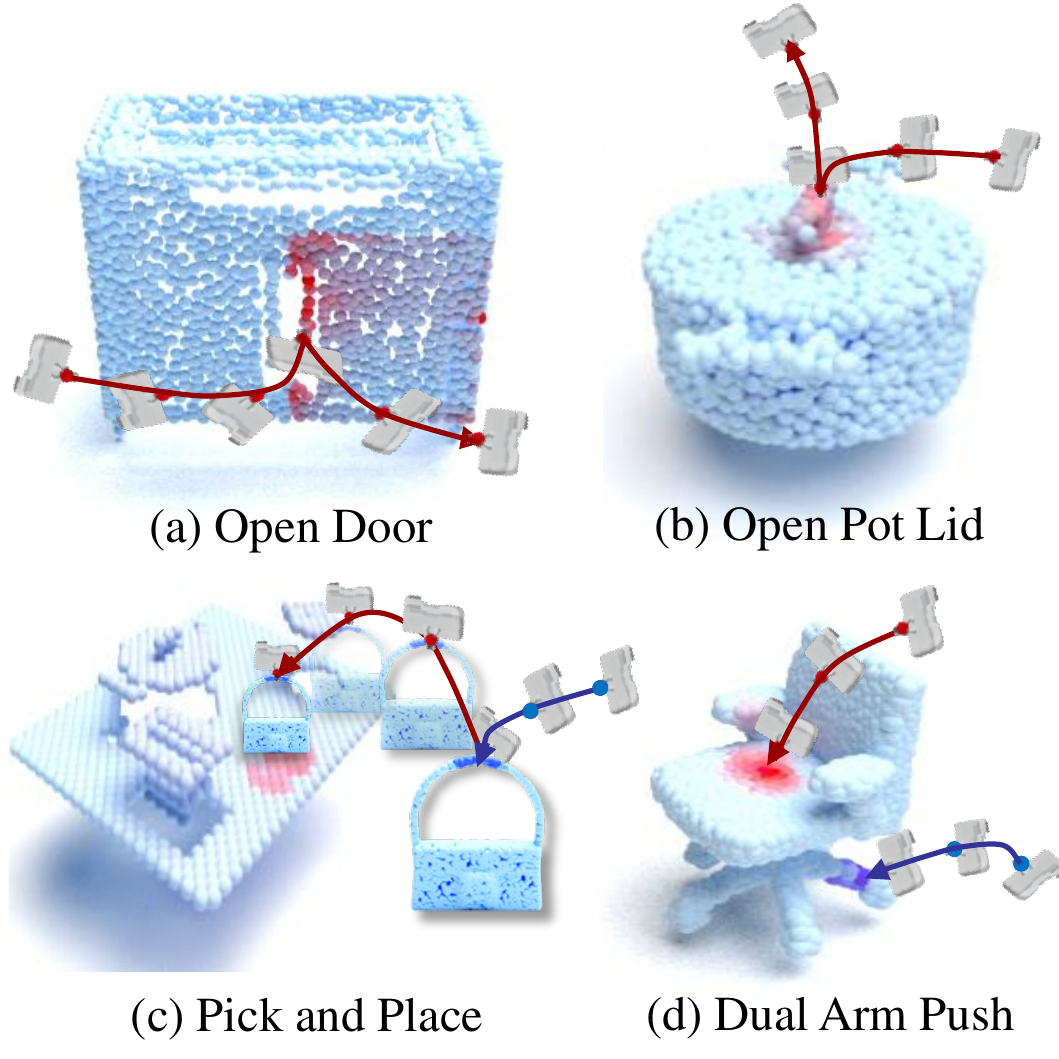}
    \caption{\textbf{Affordance examples of different manipulation tasks.} (a-b): Agent-to-object affordance map. (c): Agent-to-object and object-to-object affordance map. (d): Dual agent-to-object affordance map.}
    \label{fig:teaser}
    \vspace{-0.3cm}
\end{figure}
The existing affordance methods for manipulation have two training stages~\cite{Mo_2021_ICCV, wu2022vatmart, wang2021adaafford, zhao2022dualafford}. For example, VAT-Mart~\cite{wu2022vatmart} first trains the affordance map with data collected by an RL agent driven by curiosity, and then fine-tunes both the affordance map and the RL agent. In Where2act~\cite{Mo_2021_ICCV} and many other works~\cite{mo2021o2oafford,wang2021adaafford,zhao2022dualafford}, affordance is associated with a corresponding primitive action for each task, such as pushing and pulling.
Some recent works~\cite{borja2022affordance, wu2022learning} learn affordance through human demonstration.
A significant drawback of those two-stage methods, which first train the affordance map and then propose action sequence based on the learned affordance, is that the success rate of interaction is highly related to the accuracy of the learned affordance. Any deviations in affordance predictions will significantly reduce the task performance.

In this paper, we investigate learning affordance along with RL in an end-to-end fashion by using contact frequency to represent the affordance. Therefore the affordance is not associated with a specific primitive action but rather the contact information from past manipulation experiences of RL training.
In our method, the RL algorithm learns to utilize visual affordance generated from contact information to find the most suitable position for interaction. We also incorporate visual affordance in reward signals to encourage the RL agent to focus on points of higher likelihood.
The advantages of end-to-end affordance learning are two-fold:
 1) affordance can awaken the agent where to act as an additional observation and be incorporated into reward signals to improve the manipulation policy;
 2) learning affordance and manipulation policy simultaneously, without human demonstration or a dedicated data colleting process simplifies the learning pipeline and can migrate to other tasks easily. Additionally, it helps the affordance and manipulation policy to adapt to each other, thus producing a more robust affordance representation.
 
Using contact information as affordance evidently supports multi-stage tasks, such as picking up an object and then placing it to a proper place, as well as multi-agent tasks, such as pushing a chair with two robotic arms~\cite{borja2022affordance,lobbezoo2021reinforcement,vyas2021robotic}. These two types of tasks are difficult for two-stage affordance methods~\cite{Mo_2021_ICCV,wu2022vatmart,borja2022affordance} because they need different pre-defined data collection for each human-defined primitive action. Also, by unifying all interactions as contacts, our method can effectively represent both agent-to-object (A2O) and object-to-object (O2O) interactions, which is hard for other methods.


To test if our method can boost visual-based RL in robotic manipulation,
we conducted experiments on eight representative robot tasks, including articulated object manipulation, object pick-and-place and dual arm collaboration tasks. The results showed that our method outperformed all the baselines, including those of RL and the current two-stage affordance methods, and can successfully transfer to the real world.
To the best of our knowledge, we are the first to investigate end-to-end affordance learning for robotic manipulation. Our method can be intergrated into visual-based RL to suppport multi-stage tasks and multi-agent tasks without additional annotations or demonstrations.
    

\section{Related Work}
\label{abs}

\subsection{Robotic Manipulation Policy Learning}

The recent simulators and benchmarks have boosted the development of manipulation policy learning methods~\cite{DBLP:journals/corr/abs-2003-08515, mu2021maniskill, DBLP:journals/corr/abs-1910-10897, bi-dexhands}. For rigid object manipulation, there are already robust algorithms handling tasks such as grasping~\cite{suctionnet,DBLP:journals/corr/abs-2101-01132,borja2022affordance}, planar pushing~\cite{Li2018PushNetDP,DBLP:journals/corr/YuBFR16} and object hanging~\cite{you2021omnihang}. However, it is yet difficult to manipulate articulated objects with multiple parts despite various attempts to approach this problem from different perspectives.
For example, UMPNet~\cite{Xu2022UMPNetUM} and VAT-Mart~\cite{wu2022vatmart} utilized visual observation to directly propose action sequence, while some other studies~\cite{pmlr-v100-abbatematteo20a,DBLP:journals/corr/abs-1907-09014,eisner2022flowbot3d} achieved robust and adaptive control through model prediction. The multi-stage and multi-agent manipulation settings are also challenging for current methods~\cite{zhu2021hierarchical,mandlekar2020learning,bi-dexhands}.

\subsection{Visual Actionable Affordance Learning}
Till now, several studies have demonstrated the power of affordance representation on manipulation~\cite{Mo_2021_ICCV,wu2022vatmart}, grasping~\cite{mandikal2020graff,lenz2015deep,borja2022affordance,wu2022learning}, scene classification~\cite{dixit2015scene,zhang2014learning}, scene understanding~\cite{fowler2018human,ye2017can} and object detection~\cite{do2018affordancenet}.
The semantic information in affordance is instructive for manipulation.
Some prior affordance learning processes for manipulation, such as Where2Act~\cite{Mo_2021_ICCV}, VAT-Mart~\cite{wu2022vatmart}, AdaAfford~\cite{wang2021adaafford} and VAPO~\cite{borja2022affordance}, have two training stages. Specifically, they need to first collect interacting data to pretrain the affordance, and then train the policy based on the affordance.
For methods~\cite{https://doi.org/10.48550/arxiv.2203.00352, mandikal2020graff, nagarajan2019grounded} which train affordance and policy simultaneously, however, their affordance learning relies on human demonstration. 
Unlike them, our method requires neither pre-defined data collection process for different primitive actions\,/\,tasks nor any additional human annotations. 

\subsection{Comparison with Related Works}

The related works mentioned above studied robotic manipulation in different problem settings, the difference includes observation and annotation. 
It is hard to compare these works given their distinct settings.
Specifically,
in the door-opening task, Maniskill~\cite{mu2021maniskill} utilizes expert demonstrations for imitation learning. 
However, expert demonstrations are difficult to obtain since they are usually collected by human. 
Affordance methods, such as Where2Act~\cite{Mo_2021_ICCV}, VAT-Mart~\cite{wu2022vatmart}, and VAPO~\cite{borja2022affordance} learn the affordance prior to its policy training and are not in an end-to-end fashion. They also output gripper pose as actions which in reality have no guarantee that the gripper can reach the position. Additionally, the related affordance studies mentioned here are designed for single-stage single-agent tasks, such as opening a door or grasping an object, which has no guarantee for multi-stage or multi-agent tasks. However, using contact information for affordance natually allows RL policy to handle multi-stage tasks like picking up an object and placing it to the proper place, and multi-agent tasks like pushing a chair collaboratively by two robotic arms.

\begin{table}[t!]
    \caption{Comparison between our work and related works.}
    \begin{tabular}{c|p{0.652cm}<{\centering}|p{0.652cm}<{\centering}|p{0.652cm}<{\centering}|p{0.652cm}<{\centering}|p{0.652cm}<{\centering}|p{0.652cm}<{\centering} }
    \toprule
                    & W2A           & VAT           & MSkill    & VAPO          & Hang          & \textbf{Ours  }   \\ \hline
    No Demo         & \Checkmark    & \Checkmark    &           &               & \Checkmark    & \CheckmarkBold    \\ 
    No Full Obs     & \Checkmark    & \Checkmark    &           & \Checkmark    & \Checkmark    & \CheckmarkBold    \\ 
    End-to-End      &               &               &\Checkmark &               &               & \CheckmarkBold    \\ 
    Multi-Stage      &               &               &           &               &               & \CheckmarkBold    \\
    Multi-Agent     &               &               &\Checkmark &               &               & \CheckmarkBold    \\
    \bottomrule
    \end{tabular}
    \label{table:compare}
    \vspace{-0.3cm}
\end{table}

Table~\ref{table:compare} compares our method with five representative related works discussed above.
Listed works are Where2Act (W2A)~\cite{Mo_2021_ICCV}, VAT-Mart (VAT)~\cite{wu2022vatmart}, Maniskill (MSkill)~\cite{mu2021maniskill}, VAPO~\cite{borja2022affordance} and OmniHang (Hang)~\cite{you2021omnihang}.
"\textbf{No Demo}" means the method does not need any expert demonstrations, such as human-collected trajectories, pre-defined primitive actions and human-designed interaction poses. "\textbf{No Full Obs}" indicates the method does not need state observations of objects such as the coordinate of door handle, since the accurate state of an object is difficult to obtain in real world. 
"\textbf{End-to-End}" signifies the method trains the policy in an end-to-end fashion, \emph{i.e.}, no multiple training stages are involved and the actions of the policy can be directly applied to agents.
"\textbf{Multi-Stage}" infers the method 
can complete multi-stage tasks which the agent needs to finish multiple dependent tasks sequentially. "\textbf{Multi-Agent}" suggests the method can be adapted to multi-agent tasks where agents need to cooperate one another to finish the work.




\section{Methods}

    \begin{figure*}[t]
      \centering
      \includegraphics[trim=183.5 155 168 115,clip, scale=1.37]{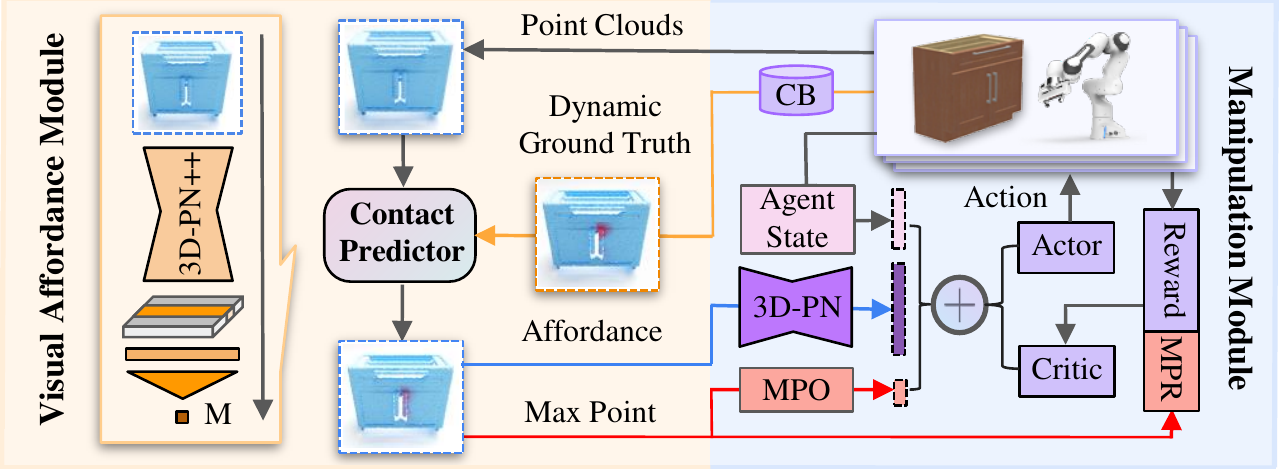}

        \caption{\textbf{Training Pipeline of End-to-End Affordance Learning.} Our pipeline contains two main modules: \textit{Manipulation Module} ($MA$ Module) generating interaction trajectories and \textit{Visual Affordance Module} ($VA$ Module) learning to generate per-point affordance map $M$ based on the real-time point cloud.
        The \textit{Contact Predictor} (CP), shared across two modules, serves as a bridge between them: 1) $MA$ Module uses the affordance map (indicated by the blue arrow) and \textit{Max-affordance Point Observation} ($MPO$) (indicated by the upper red arrow) predicted by the CP as a part of the input observation. A \textit{Max-affordance Point Reward} ($MPR$) feedback (indicated by the lower red arrow) is also incorporated in training $MA$ Module; 2) $MA$ Module maintains a $\textit{Contact Buffer}$ ($CB$) by collecting collision information and generating \textit{Dynamic Ground Truth} ($
       DGT$) (indicated by the orange arrow), where $VA$ Module uses the $DGT$ as the target for training $CP$.}
      \label{fig:pipeline}
      \vspace{-0.5cm}
    \end{figure*}

\subsection{Method Overview}

Visual-based RL is increasingly valued on robotic manipulation tasks, especially those requiring the agent to manipulate different objects with a single policy. Meanwhile, recent studies~\cite{srinivas2020curl,stooke2021decoupling,wu2022learning} identified the difficulty of 
learning observation encoders by RL from high-dimensional inputs such as point clouds and images. In our framework, we tackled this critical problem by exploiting underlying information through a process called "Contact Prediction".

In manipulation settings, contact is the fundamental way humans interact with an object. We believe that physical contact positions during interactions reflect the understanding of crucial semantic information about the object (\emph{e.g.}, a human grasp a handle to open a door because the handle provides the position to apply force). 

We proposed a novel end-to-end RL learning framework for manipulating 3D objects. As shown in Fig.~\ref{fig:pipeline}, our framework is comprised of two parts. 1) $\textit{Manipulation Module}$ ($MA$ Module) is a RL framework which uses the affordance map predicted by a \textit{Contact Predictor} ($CP$) as an additional observation and reward signal; 2) $\textit{Visual Affordance Module}$ ($VA$ Module) is a per-point scoring network, which uses
the contact positions collected from RL training process as the \textit{Dynamic Ground Truth} ($DGT$) to indicate the position of interaction.




Concretely, at every time-step $t$, the $MA$ Module 
outputs an action $a_t$
based on
the robotic arm state $s_t$ (\emph{i.e.,} the angle and angular velocity of each joint) and the affordance map $M_t$ predicted by the $VA$ Module. 
After each time-step $t$, the contact position in RL training is inserted to the $\textit{Contact Buffer}$ ($CB$). 
After each $k$ time-steps, we integrate the data in $CB$ to generate the per-point score as the $DGT$ to update the $VA$ Module. 

\begin{algorithm}[t]
\caption{End-to-End Affordance Learning.}\label{alg:DERL_with_arbitrary_policies}
\begin{algorithmic}
\Require $E$: the environment, $CB$: current contact buffer, $RL$: RL pipeline, $i$: current timestep.
\Ensure $a$: an action generated by RL pipeline
\State $c \gets collectContact(E, RL)$

\State $CB \gets insert(CB, c)$

\State $PC \gets getPointCloud(E)$\Comment{Point Cloud}

\If { $i \% k = 0$}\Comment{Update CP every k timesteps}
    \State $DGT \gets getMap(PC, CB)$\Comment{Dynamic Ground Truth}
    
    \State $CP \gets update(CP, DGT)$\Comment{Update CP network}
\EndIf

\State $M \gets CP(PC)$\Comment{Affordance map}

\State $RL \gets train(RL, E, M)$\Comment{Update RL network}

\State $a \gets RL(E, M)$\Comment{The action}

\State \textbf{return} $a$
\end{algorithmic}
\end{algorithm}

\subsection{Visual Affordance Module: \emph{Contact as Prior}}


During robotic manipulation, physical contacts naturally happen between agent and object, or object and object. As contacts do not relate to any human-defined primitive action such as pull or push, the contact position is a general representation, providing visual prior for manipulation.

The RL training pipeline in Manipulation Module ($MA$ Module) continuously interacts with the environment to collect 1) the partial point cloud observation $\mathcal{P}$, 2) the contact position under object coordinate.
Based on this information, we measure how likely a contact between agent and object (A2O) or between object and object (O2O) is going to happen by the per-point contact \textit{frequency} as the affordance during the current RL training.
The Visual Affordance Module ($VA$ Module) then learn to predict the per-point \textit{frequency}. The training details of $VA$ Module is as follow.

\textbf{Input:}
Following the prior studies~\cite{wang2021adaafford, wu2022vatmart, Mo_2021_ICCV}, the input for $VA$ Module contains a partial point cloud observation $\mathcal{P}$.

\textbf{Output:}
The output of the $VA$ Module is a per-point affordance map $\textit{M}$ for each of the point from the input. The map contains A2O affordance and O2O affordance. 

\textbf{Network Architecture:}
The prediction is completed by a \textit{Contact Predictor} ($CP$) that uses a PointNet++ \cite{qi2017pointnetplusplus} to extract a per-point feature $f\in \mathbb{R}^{128}$ from point cloud observation $\mathcal{P}$, the feature $f$ is then fed through a Multi-Layer Perceptron (MLP) to predict the per-point actionable affordance~\cite{Mo_2021_ICCV}.

\textbf{Dynamic Ground Truth:}
To connect the RL pipeline in $MA$ Module with the $VA$ Module, we use a \textit{Contact Buffer} $CB$ to keep $l$ record of history contact points, and to compute the $DGT$.
Specifically, each object in the training set has a corresponding $CB$, it records contact positions on the object. To maintain the buffer size, the buffer randomly evicts one record whenever a new record of contact event is inserted.
To provide training ground truth for $CP$, we calculate the $DGT$ by first calculating the number of contacts within radius $r$ from each point on the object point cloud, and then applying normalization to obtain \textit{Dynamic Ground Truth} $DGT$.
The normalization is as follow:
\begin{equation}
    DGT_t^i(p)=\frac{\sum_{q\in CB_t^i}I(|p-q|_2<r)}{\max_{p'}\sum_{q\in CB_t^i}I(|p'-q|_2<r) + \epsilon} ,
\end{equation}
where $DGT_t^i$ indicates the \textit{Dynamic Ground Truth} for object $i$ at time-step $t$, $CB_t^i$ is the corresponding \textit{Contact Buffer}.

\textbf{Training:}
The $CP$ is updated with $DGT^i_t$ as below:
\begin{equation}
    \textit{CP}_t^*=\argmin_{\textit{CP}}\sum_isr_t^i\left|\left|\sum_{p\in \mathcal{P}^i}\textit{CP}(p|\mathcal{P}^i)-DGT_t^i(p)\right|\right|_2
\end{equation}
where $sr_t^i$ is the current manipulation success rate on object $i$, $\mathcal{P}^i$ is the pointcloud of $i$-th object and $\textit{CP}_t^*$ is the optimal $\textit{CP}$ .

\subsection{Manipulation Module: \emph{Affordance as Guidance}}


\textit{Manipulation Module} ($MA$ Module) is an RL framework able to learn to manipulate objects from scratch. Different from previous methods~\cite{wu2022vatmart, Mo_2021_ICCV,mu2021maniskill}, our $MA$ Module takes advantage of both the reward and observation generated by the $VA$ Module.

\textbf{Input:} The input for $MA$ Module includes, 1) a point cloud $\mathcal{P}$ of the real-time environment~\cite{Mo_2021_ICCV,wu2022vatmart}; 2) an affordance map $M$ generated by $VA$ Module; 3) the state $s$ of the robotic arm.  The state $s$ consists of position, velocity and angle of each joint of the robotic arm; 4) a state-based \textit{Max-affordance Point Observation} ($MPO$), which indicates the point with the maximum affordance score on $\mathcal{P}$ .

\textbf{Output:} The output of the $MA$ Module is an action $a$, which is then executed by the robotic arm. In our setting, the RL policy controls each joint of the robotic arm directly.

\textbf{Reward from Affordance:} We introduce the \textit{Max-affordance Point Reward} ($MPR$) into our pipeline, where a point on the point cloud with maximum affordance score predicted by the $VA$ Module is selected as the guidance for learning $MA$ Module. We use the distance between robot end-effector and this selected point to compute an additional reward in the RL process. We found this reward from affordance could benefit the RL training thus improve the overall performance.

\textbf{Network Architecture:} 
The policy of the $MA$ Module is a neural network $\pi_\theta$ with learnable parameter $\theta$. The network consists of a PointNet \cite{DBLP:journals/corr/QiSMG16} and a MLP. The PointNet extracts feature $f\in \mathbb{R}^{128}$ from the point cloud $\mathcal{P}$, affordance map $M$ and additional masks $m$. The extracted feature $f$ is then concatenated with $s$ and fed to the MLP to obtain actions.

\textbf{Training:}
We use Proximal Policy Optimization (PPO) algorithm~\cite{DBLP:journals/corr/SchulmanWDRK17} to train the $MA$ Module.
To improve the training efficiency by exploiting the high parallelism of our simulator, we deploy $k$ different objects in the simulator, each object is replicated $n$ times and given to one or two robotic arms. Hence, there are a total of $k \times n$ environments, each with a robotic arm (or two robotic arms in our multi-agent tasks) interacting with an object, as shown in Fig.~\ref{fig:map}. 

\section{Experiment}
\label{exp}
\subsection{Task Description}
To evaluate our method, we designed three types of manipulation tasks: single-stage, multi-stage and multi-agent.
In all tasks, a robotic arm or two robotic arms are required to complete a specific manipulation task on different objects.

The first type of tasks are single-stage manipulation tasks as follow:

\textbf{Close Door:} 
A door is initially open to a specific angle.
The agent need to close the door completely. We increase the difficulty of this task by applying an additional force on the door attempting to keep the door to the initial position and doubling the friction of the hinge. 

\textbf{Open Door:} 
A door is initially closed. 
The agent need to open the door to a specific angle. 
This task can test whether the agent learn to leverage key parts like the handle to open the door, which is challenging.

\textbf{Push Drawer:} 
A drawer is initially open to a specific distance.
similar to \textit{close door}, the agent need to close the drawer on a cabinet completely.

\textbf{Pull Drawer:} A drawer is initially closed, similar to \textit{open door}, the agent need to open the drawer to a specific distance.

\textbf{Push Stapler:} A stapler is on the desk, initially open. The agent need to push on the stapler and close it.

\textbf{Lift Pot Lid:} A pot is on the floor with its lid on. The agent need to lift the lid.

To show the agent can learn a policy in a multi-stage task, we use the pick-and-place task as follow:

\textbf{Pick and Place:} 
An object should be picked up and then placed on a table that already have several random objects on it, both the table and objects are randomly selected from the given datasets. The agent need to place the object stably on the table without collision.

To show our method can be generalized to multi-agent settings, we use the dual-arm-push task as follow:

\textbf{Dual Arm Push:}
Two robotic arms need to be controled to push a chair to a specific distance and prevent the chair from falling over.

To make the agent better adapt to the environment, we add a movable base to the arm, allowing the arm to move horizontally within a specific range. 
The reward designs and other details are listed on our website.

\begin{figure*}[t]
  \centering
\includegraphics[trim=52 39 10 5,clip, scale=0.54]{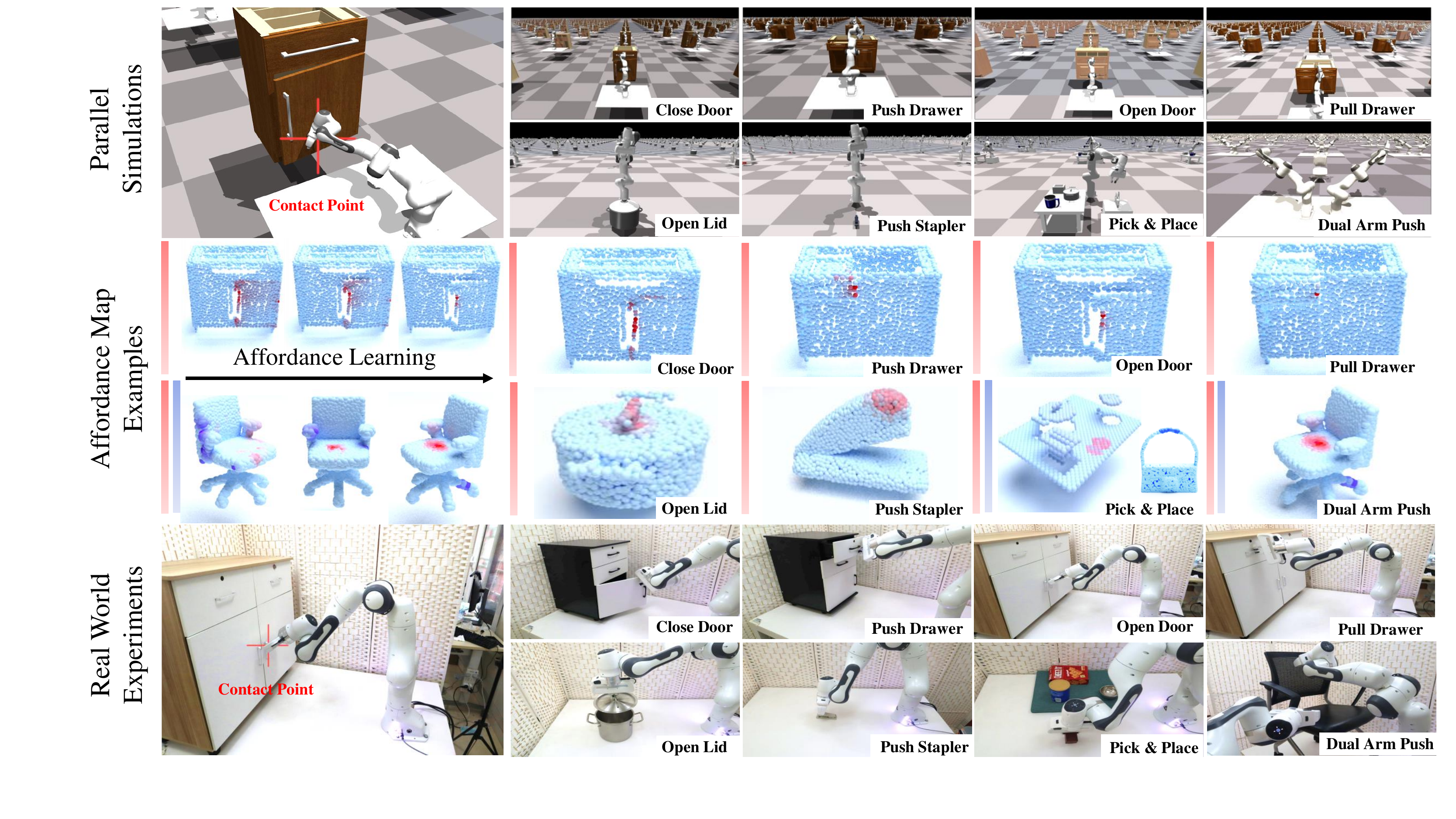}
  \caption{\textbf{Experiment Settings and Affordance Learning Visualization.} Top: the tasks settings in simulators. Middle: the change in affordance maps during end-to-end training and the final affordance map examples. Bottom: the real-world experiments.}
  \label{fig:map}
\end{figure*}
\subsection{Dataset and Simulator}
\label{dataset}
We performed our experiments using the Isaac Gym simulator~\cite{DBLP:journals/corr/abs-2108-10470}.
We used Franka Panda robot arm as the agent for all tasks.
Our training and testing data are the subset of the PartNet-Mobility dataset~\cite{chang2015shapenet} and VAPO dataset~\cite{borja2022affordance}. 
For tasks \textbf{Close Door} and \textbf{Open Door}, we divided the objects with door handles in the StorageFurniture category into four subcategories: \textit{one door left}, \textit{one door right}, \textit{two door left} and \textit{two door right}.
For tasks \textbf{Pull Drawer} and \textbf{Push Drawer}, we divided the objects with door handles in the StorageFurniture category into two subcategories: \textit{drawer without door} and \textit{drawer with door}.
For tasks \textbf{Push Stapler} and \textbf{Lift Pot Lid}, we chose all \textit{Stapler} and \textit{Pot} from PartNet-Mobility dataset.
For task \textbf{Pick and Place}, we chose three representative categories of objects from VAPO dataset to pick. We also selected four types of different tables: \textit{Round Table}, \textit{Triangle Table}, \textit{Square Table} and \textit{Irregular Table} and three daily items from PartNet-Mobility dataset were placed randomly on the table.
For task \textbf{Dual Arm Push}, we chose 60 \textit{Chairs} from PartNet-Mobility dataset.

\subsection{Baselines and Ablations}

We compared our method with seven baselines:
\begin{itemize}
    \item Where2act~\cite{Mo_2021_ICCV}: the original method only generates single-stage interaction proposals. To use this method as a baseline in our tasks, we implemented a multi-stage Where2act baseline (up to six steps). The object is gradually altered by pushing or pulling interactions produced by Where2act until the task is completed or the maximum number of steps have been taken. Unlike our own setting, this baseline used a flying gripper instead of a robotic arm.
    \item VAT-Mart~\cite{wu2022vatmart}: 
    We followed the implementation of paper~\cite{wu2022vatmart}. Similar to Where2act, we implemented this method in our environment as a baseline with a flying gripper.
    \item RL: we used a point cloud based PPO as our baseline. 
    \item RL+Where2act: we replaced the Contact Predictor in our method with a pre-trained Where2act model that can output a per-point actionable score. 
    The parameters in the Where2act model is frozen when training $MA$ Module.
    \item RL+O2OAfford and RL+O2OAfford+Where2act: Similar to RL+Where2act, we replaced the O2O affordance map in our method with the map produced by a pre-trained O2OAfford~\cite{mo2021o2oafford} model.
    \item MAPPO: we used a point cloud based multi-agent RL algorithm (MARL): MAPPO~\cite{yu2021surprising} as our baseline.
    \item Multi-Task RL~\cite{DBLP:journals/corr/SchulmanWDRK17}: we adapted PPO to the multi-task setting by providing the one-hot task ID as input. To make this method comparable on the test set, both the test set and the training set were used in training process. So this is an oracle baseline.
\end{itemize}

\begin{table*}[]
\centering
\setlength\tabcolsep{5.4pt}
\caption{\textbf{
Quantitative results of single-stage tasks. (More results on our website.)}}
\label{table:ablation}
\label{table:asr}
\begin{tabular}{c|cccc|cccc|cccc|cccc}
\toprule
 \multirow{3}{*}{\diagbox{Methods}{Datasets}}       & \multicolumn{4}{c|}{Open Door}  & \multicolumn{4}{c|}{Pull Drawer}      & \multicolumn{4}{c|}{Push Stapler}   & \multicolumn{4}{c}{Open Pot Lid}     \\ \cline{2-17} 
             & \multicolumn{2}{c|}{ASR}          & \multicolumn{2}{c|}{MP} & \multicolumn{2}{c|}{ASR}          & \multicolumn{2}{c|}{MP} & \multicolumn{2}{c|}{ASR}          & \multicolumn{2}{c|}{MP} & \multicolumn{2}{c|}{ASR}          & \multicolumn{2}{c}{MP} \\
             & train & \multicolumn{1}{c|}{test} & train       & test      & train & \multicolumn{1}{c|}{test} & train       & test      & train & \multicolumn{1}{c|}{test} & train       & test      & train & \multicolumn{1}{c|}{test} & train      & test      \\ \hline
Where2act     &22.8           & \multicolumn{1}{c|}{14.1}          &6.8             &8.3             &19.0     & \multicolumn{1}{c|}{12.9}   &2.3            &0.0            &  16.4             &  \multicolumn{1}{c|}{14.4}             &    13.0             &      13.0          &  10.5               & \multicolumn{1}{c|}{5.4}              &  8.7           &      4.3         \\
VAT-Mart      &23.2           & \multicolumn{1}{c|}{21.9}          &31.8            &33.3            &5.5      & \multicolumn{1}{c|}{5.1}    &0.0           &0.0            &         21.9      &  \multicolumn{1}{c|}{20.9}             &   17.4              &      13.0          &        27.4         & \multicolumn{1}{c|}{21.5}              &    17.4         &  17.4\\
Multi-task RL                 & 18.8                      & \multicolumn{1}{c|}{9.2}& 11.4          & 5.0          & 0.1             & \multicolumn{1}{c|}{2.4}           & 0.0            & 2.8        &   34.9   & \multicolumn{1}{c|}{30.2}          &  30.4     &  26.1 & 35.2 & \multicolumn{1}{c|}{32.6} &21.7 &17.4 \\
RL             & 21.5         & \multicolumn{1}{c|}{5.5}& 22.7          & 0.0         & 23.1            & \multicolumn{1}{c|}{22.4}          & 19.6           & 19.5         &45.5    & \multicolumn{1}{c|}{40.6} 
& 34.8         & 30.4 &  32.5 & \multicolumn{1}{c|}{28.6}       &    21.7  & 21.7  \\
RL+Where2act             & 20.5         & \multicolumn{1}{c|}{8.0}& 19.3          & 9.4         & 25.2            & \multicolumn{1}{c|}{22.2}          & 24.4           & 21.9         &48.9    & \multicolumn{1}{c|}{45.2} 
& 39.1     & 34.8 &  38.2 & \multicolumn{1}{c|}{30.6}       &    26.1  & 21.7  \\\hdashline[1pt/1pt]

\textbf{Ours}& \textbf{52.9}             & \multicolumn{1}{c|}{\textbf{32.6}} & \textbf{61.4} & \textbf{41.7}  & 59.7 & \multicolumn{1}{c|}{\textbf{58.6}} & 62.8 &\textbf{63.3}         & \textbf{69.5} & \multicolumn{1}{c|}{\textbf{53.2}} & \textbf{47.8} & \textbf{39.1}  &   \textbf{49.5} & \multicolumn{1}{c|}{\textbf{44.6}}          & \textbf{34.8} & \textbf{30.4}  \\ 
\hdashline[1pt/1pt]

Ours w/o MPO   & 48.0           & \multicolumn{1}{l|}{23.8} & 50.0                    & 16.7                 & 41.9                             & \multicolumn{1}{l|}{42.5}          & 38.6           & 43.8      &   60.6   & \multicolumn{1}{l|}{52.5}           &  43.5 & \textbf{39.1}  & 44.2   & \multicolumn{1}{l|}{40.7} &      \textbf{34.8}       &\textbf{30.4} \\
Ours w/o MPR   &28.2         & \multicolumn{1}{l|}{8.4}  &29.5                     &8.3                   &\textbf{62.3}                     & \multicolumn{1}{l|}{44.0}         &\textbf{65.9}    &43.8         &   50.8   & \multicolumn{1}{l|}{39.9}            &39.1  & 30.4  & 44.8      & \multicolumn{1}{l|}{40.1}     & 30.4         &  26.1\\
Ours w/o E2E   & 21.2           & \multicolumn{1}{l|}{12.4} &20.5                     &8.3                     & 57.7                              & \multicolumn{1}{l|}{57.3}           & 61.1            & 61.7        &   40.2   & \multicolumn{1}{l|}{36.6}           &  39.1 & 34.8 & 32.1   &\multicolumn{1}{l|}{30.6} &      30.4         & 26.1\\

\bottomrule
\end{tabular}
\end{table*}

\begin{table*}[]
    \begin{minipage}[t]{.49\linewidth}
        \centering
        \caption{\textbf{Quantitative results of Pick-and-Place.}}
        \vspace{-0.2cm}
        \label{table:pap-asrmr}
        \label{table:pap-ablation}
        \begin{tabular}{c|cc|cc} 
        \toprule
        \multirow{2}{*}{\diagbox{Methods}{Metrics}} & \multicolumn{2}{c|}{ASR} & \multicolumn{2}{c }{MP} \\
                                & train        & test         & train       & test          \\ 
        \hline
        RL                     &25.2           & 22.1        & 19.2     & 11.5          \\
        RL+O2OAfford           & 26.1          & 22.2        & 19.2     & 11.5             \\
        RL+Where2act           & 28.6          & 23.5        & 23.1     & 15.4           \\
        RL+O2OAfford+Where2act & 30.5          & 26.2        & 23.1    & 15.4        \\ \hdashline[1pt/1pt]
        
        \textbf{Ours}           &\textbf{46.5} &\textbf{39.2} &\textbf{30.7}&\textbf{26.9}  \\ \hdashline[1pt/1pt]
        Ours w/o A2O Map        &26.7          &22.3          &23.1        &19.2           \\
        Ours w/o O2O Map        &31.9          &26.2          &23.1         &15.4          \\
        Ours w/o MPO            & 40.1         &  30.2        & 19.2        & 15.4   \\
        Ours w/o MPR            &  36.2        &    33.5      & \textbf{30.7}        &  23.1  \\
        Ours w/o E2E            &  30.2        &    21.4      & 26.9        &  19.2  \\

        \bottomrule
        \end{tabular}
    \end{minipage}
    \begin{minipage}[t]{.49\linewidth}
        \centering
        \caption{\textbf{Quantitative results of dual-arm-push.}}
        \vspace{-0.2cm}
        \label{table:dap-asrmr}
        \label{table:dap-ablation}
        \begin{tabular}{c|cc|cc} 
        \toprule
        \multirow{2}{*}{\diagbox{Methods}{Metrics}} & \multicolumn{2}{c|}{ASR} & \multicolumn{2}{c }{MP} \\
                                & train        & test         & train        & test           \\ 
        \hline
        MAPPO                   &7.8           &9.0           &0.0           &0.0             \\
        RL                      &37.2          &36.1          &36.4          &31.3            \\
        Multi-task RL          &51.6          &52.9          &54.5          &56.3            \\ \hdashline[1pt/1pt]

        \textbf{Ours}           &83.9          &78.5          &90.9          &93.8            \\ \hdashline[1pt/1pt]
        Ours w/o MPO            &\textbf{95.9} &\textbf{96.3} &\textbf{100.0}&\textbf{100.0}  \\
        Ours w/o MPR            &63.9          &55.3          &63.6          &56.3            \\
        Ours w/o E2E            &53.5          &55.9          &56.8          &50.0            \\
        \bottomrule
        \end{tabular}
    \end{minipage}
    \vspace{-0.5cm}
\end{table*}

To further evaluate the importance of different components of our method, we conducted ablation study by comparing our method with five ablations:
\begin{itemize}
    \item Ours w/o MPR: ours without the max-point reward.
    \item Ours w/o MPO: ours without the max-point obversation.
    \item Ours w/o E2E: our method trained by a two-stage procedure. The $VA$ Module is trained upon a fixed pretrained $MA$ Module. The $MA$ Module is then fine-tuned on the freezed $VA$ Module. 
    \item Ours w/o A2O Map: our method without agent-to-object affordance map in multi-stage tasks.
    \item Ours w/o O2O Map: our method without object-to-object affordance map in multi-stage tasks.
\end{itemize}

\subsection{Evaluation Metrics}


For each task, we trained the method (ours, baselines and ablations) on the training set and saved checkpoints every 3200 time-steps within $160,000$ total time-steps. After training, we chose the checkpoint with the largest average success rate on training set for comparison, the method was tested on eight different random seeds.
We adpoted two metrics to measure the performance: 
\begin{itemize}
\item Average Success Rate (ASR): The ASR 
is the average of the algorithm's success rate on all objects in the training\,/\,testing dataset.
\item Master Percentage (MP): We assume a policy is ``stable'' on an object if it has a success rate of more than 50\% on that object. The master percentage 
is the percentage of objects which the algorithm can success with a probability greater than or equal to $50\%$. If an algorithm manages to reach a success rate over $50\%$ on a certain object, it is expected to success within two trials.
\end{itemize}

\textbf{Due to the page limit, we listed the results of Close Door and Push Drawer and the variance of the reported metric on our website.}

\subsection{Baseline Comparision and Ablation Study}
From~\Cref{table:ablation,table:pap-ablation}, the results of Where2act and RL show the visual affordance can improve the RL performance. However, our method achieves a more significant improvement over baselines in both training and testing sets. 
In dual-arm-push, as Table~\ref{table:dap-ablation} shows, our method outperforms both RL and MARL methods.

From all tables, we see the MPO, MPR and E2E components play important roles in our method except that E2E on dual-arm-push. The potential reason is that the predicted max affordance point on the object is changing during object movement, which may influence the RL training. This may be something worth looking into in the future.

Fig.~\ref{fig:map} shows the change in affordance maps during end-to-end training and examples of final affordance maps. We can see that as the training proceeds, the affordance map gradually concentrates. More qualitative results can be found on our website.

\subsection{Real-world Experiment}
We used a digital twin system~\cite{XIA2021210} for real-world experiment: The training process was in simulation, we then used some unseen objects to evaluate our method in real world. The input of the agent has two folds: 1) point cloud input from simulator, 2) agent state input from real world.
The actions of the agent were computed upon the combination of the two input sources, and then were applied to the robotic arms both in the simulator and the real world.
The experiment settings are shown in Fig.\ref{fig:map}.
Experiments show that our trained model can successfully transfer to the real world. The video and more details can be found on our website \url{https://sites.google.com/view/rlafford/}.

\section{Conclusion}
\label{conc}

To the best of our knowledge, this the first work that  proposes an end-to-end affordance RL framework for robotic manipulation tasks. 
In RL training, affordance can improve the policy learning by providing additional observation and  reward signals. Our framework automatically learns affordance semantics 
through RL training without human demonstration or other artificial  designs dedicated to data collection. 
The simplicity of our method, together  with the superior performance over strong baselines and the wide range of applicable tasks, has demonstrated  the effectiveness of learning from contact information. We believe our work could potentially open a new way for future RL-based manipulation developments.

\section*{ACKNOWLEDGEMENT}This project was supported by the National Natural Science Foundation of China (No. 62136001). We would like to thank Hongchen Wang, Ruihai Wu, Yan Zhao and Yicheng Qian for the helpful discussion and baseline implementation, and Ruimin Jia for suggestions in paper writing.



{
\bibliographystyle{IEEEtran}
\bibliography{IEEEabrv,reference}
}

\end{document}